# Variational Probabilistic Inference
# and the QMR-DT Network

**Tommi S. Jaakkola**                                    TOMMI@AI.MIT.EDU
*Artificial Intelligence Laboratory,*
*Massachusetts Institute of Technology,*
*Cambridge, MA 02139 USA*

**Michael I. Jordan**                                    JORDAN@CS.BERKELEY.EDU
*Computer Science Division and Department of Statistics,*
*University of California,*
*Berkeley, CA 94720-1776 USA*

## Abstract

We describe a variational approximation method for efficient inference in large-scale probabilistic models. Variational methods are deterministic procedures that provide approximations to marginal and conditional probabilities of interest. They provide alternatives to approximate inference methods based on stochastic sampling or search. We describe a variational approach to the problem of diagnostic inference in the "Quick Medical Reference" (QMR) network. The QMR network is a large-scale probabilistic graphical model built on statistical and expert knowledge. Exact probabilistic inference is infeasible in this model for all but a small set of cases. We evaluate our variational inference algorithm on a large set of diagnostic test cases, comparing the algorithm to a state-of-the-art stochastic sampling method.

## 1. Introduction

Probabilistic models have become increasingly prevalent in AI in recent years. Beyond the significant representational advantages of probability theory, including guarantees of consistency and a naturalness at combining diverse sources of knowledge (Pearl, 1988), the discovery of general exact inference algorithms has been principally responsible for the rapid growth in probabilistic AI (see, e.g., Lauritzen & Spiegelhalter, 1988; Pearl, 1988; Shenoy, 1992). These exact inference methods greatly expand the range of models that can be treated within the probabilistic framework and provide a unifying perspective on the general problem of probabilistic computation in graphical models.

Probability theory can be viewed as a combinatorial calculus that instructs us in how to merge the probabilities of sets of events into probabilities of composites. The key operation is that of marginalization, which involves summing (or integrating) over the values of variables. Exact inference algorithms essentially find ways to perform as few sums as possible during marginalization operations. In terms of the graphical representation of probability distributions—in which random variables correspond to nodes and conditional independencies are expressed as missing edges between nodes—exact inference algorithms define a notion of "locality" (for example as cliques in an appropriately defined graph), and attempt to restrict summation operators to locally defined sets of nodes.

While this approach manages to stave off the exponential explosion of exact probabilistic computation, such an exponential explosion is inevitable for any calculus that explicitly performs summations over sets of nodes. That is, there are models of interest in which "local" is overly large (see Jordan, et al., in press). From this point of view, it is perhaps not surprising that exact inference is NP-hard (Cooper, 1990).

In this paper we discuss the inference problem for a particular large-scale graphical model, the Quick Medical Reference (QMR) model.[1] The QMR model consists of a combination of statistical and expert knowledge for approximately 600 significant diseases and approximately 4000 findings. In the probabilistic formulation of the model (the QMR-DT), the diseases and the findings are arranged in a bi-partite graph, and the diagnosis problem is to infer a probability distribution for the diseases given a subset of findings. Given that each finding is generally relevant to a wide variety of diseases, the graph underlying the QMR-DT is dense, reflecting high-order stochastic dependencies. The computational complexity of treating these dependencies exactly can be characterized in terms of the size of the maximal clique of the "moralized" graph (see, e.g., Dechter, 1998; Lauritzen & Spiegelhalter, 1988). In particular, the running time is exponential in this measure of size. For the QMR-DT, considering the standardized "clinocopathologic conference" (CPC) cases that we discuss below, we find that the median size of the maximal clique of the moralized graph is 151.5 nodes. This rules out the use of general exact algorithms for the QMR-DT.

The general algorithms do not take advantage of the particular parametric form of the probability distributions at the nodes of the graph, and it is conceivable that additional factorizations might be found that take advantage of the particular choice made by the QMR-DT. Such a factorization was in fact found by Heckerman (1989); his "Quickscore algorithm" provides an exact inference algorithm that is tailored to the QMR-DT. Unfortunately, however, the run time of the algorithm is still exponential in the number of positive findings. For the CPC cases, we estimate that the algorithm would require an average of 50 years to solve the inference problem on current computers.

Faced with the apparent infeasibility of exact inference for large-scale models such as the QMR-DT, many researchers have investigated approximation methods. One general approach to developing approximate algorithms is to perform exact inference, but to do so partially. One can consider partial sets of node instantiations, partial sets of hypotheses, and partial sets of nodes. This point of view has led to the development of algorithms for approximate inference based on heuristic search. Another approach to developing approximation algorithms is to exploit averaging phenomena in dense graphs. In particular, laws of large numbers tell us that sums of random variables can behave simply, converging to predictable numerical results. Thus, there may be no need to perform sums explicitly, either exactly or partially. This point of view leads to the variational approach to approximate inference. Finally, yet another approach to approximate inference is based on stochastic sampling. One can sample from simplified distributions and in so doing obtain information about a more complex distribution of interest. We discuss each of these methods in turn.

Horvitz, Suermondt and Cooper (1991) have developed a partial evaluation algorithm known as "bounded conditioning" that works by considering partial sets of node instan-tiations. The algorithm is based on the notion of a "cutset"; a subset of nodes whose removal renders the remaining graph singly-connected. Efficient exact algorithms exist for singly-connected graphs (Pearl, 1988). Summing over all instantiations of the cutset, one can calculate posterior probabilities for general graphs using the efficient algorithm as a subroutine. Unfortunately, however, there are exponentially many such cutset instantiations. The bounded conditioning algorithm aims at forestalling this exponential growth by considering partial sets of instantiations. Although this algorithm has promise for graphs that are "nearly singly-connected," it seems unlikely to provide a solution for dense graphs such as the QMR-DT. In particular, the median cutset size for the QMR-DT across the CPC cases is 106.5, yielding an unmanageably large number of $2^{106.5}$ cutset instantiations.

---

1. The acronym "QMR-DT" that we use in this paper refers to the "decision-theoretic" reformulation of the QMR by Shwe, et al. (1991). Shwe, et al. replaced the heuristic representation employed in the original QMR model (Miller, Fasarie, & Myers, 1986) by a probabilistic representation.

Another approach to approximate inference is provided by "search-based" methods, which consider node instantiations across the entire graph (Cooper, 1985; Henrion, 1991; Peng & Reggia, 1987). The general hope in these methods is that a relatively small fraction of the (exponentially many) node instantiations contains a majority of the probability mass, and that by exploring the high probability instantiations (and bounding the unexplored probability mass) one can obtain reasonable bounds on posterior probabilities. The QMR-DT search space is huge, containing approximately $2^{600}$ disease hypotheses. If, however, one only considers cases with a small number of diseases, and if the hypotheses involving a small number of diseases contain most of the high probability posteriors, then it may be possible to search a significant fraction of the relevant portions of the hypothesis space. Henrion (1991) was in fact able to run a search-based algorithm on the QMR-DT inference problem, for a set of cases characterized by a small number of diseases. These were cases, however, for which the exact Quickscore algorithm is efficient. The more general corpus of CPC cases that we discuss in the current paper is not characterized by a small number of diseases per case. In general, even if we impose the assumption that patients have a limited number $N$ of diseases, we cannot assume a priori that the model will show a sharp cutoff in posterior probability after disease $N$. Finally, in high-dimensional search problems it is often necessary to allow paths that are not limited to the target hypothesis subspace; in particular, one would like to be able to arrive at a hypothesis containing few diseases by pruning hypotheses containing additional diseases (Peng & Reggia, 1987). Imposing such a limitation can lead to failure of the search.

More recent partial evaluation methods include the "localized partial evaluation" method of Draper and Hanks (1994), the "incremental SPI" algorithm of D'Ambrosio (1993), the "probabilistic partial evaluation" method of Poole (1997), and the "mini-buckets" algorithm of Dechter (1997). The former algorithm considers partial sets of nodes, and the latter three consider partial evaluations of the sums that emerge during an exact inference run. These are all promising methods, but like the other partial evaluation methods it is yet not clear if they restrict the exponential growth in complexity in ways that yield realistic accuracy/time tradeoffs in large-scale models such as the QMR-DT.[2]

Variational methods provide an alternative approach to approximate inference. They are similar in spirit to partial evaluation methods (in particular the incremental SPI and mini-buckets algorithms), in that they aim to avoid performing sums over exponentially many summands, but they come at the problem from a different point of view. From the variational point of view, a sum can be avoided if it contains a sufficient number of terms such that a law of large numbers can be invoked. A variational approach to inference replaces quantities that can be expected to be the beneficiary of such an averaging process with surrogates known as "variational parameters." The inference algorithm manipulates these parameters directly in order to find a good approximation to a marginal probability of interest. The QMR-DT model turns out to be a particularly appealing architecture for the development of variational methods. As we will show, variational methods have a simple graphical interpretation in the case of the QMR-DT.

---

2. D'Ambrosio (1994) reports "mixed" results using incremental SPI on the QMR-DT, for a somewhat more difficult set of cases than Heckerman (1989) and Henrion (1991), but still with a restricted number of positive findings.

A final class of methods for performing approximate inference are the stochastic sampling methods. Stochastic sampling is a large family, including techniques such as rejection sampling, importance sampling, and Markov chain Monte Carlo methods (MacKay, 1998). Many of these methods have been applied to the problem of approximate probabilistic inference for graphical models and analytic results are available (Dagum & Horvitz, 1993). In particular, Shwe and Cooper (1991) proposed a stochastic sampling method known as "likelihood-weighted sampling" for the QMR-DT model. Their results are the most promising results to date for inference for the QMR-DT—they were able to produce reasonably accurate approximations in reasonable time for two of the difficult CPC cases. We consider the Shwe and Cooper algorithm later in this paper; in particular we compare the algorithm empirically to our variational algorithm across the entire corpus of CPC cases.

Although it is important to compare approximation methods, it should be emphasized at the outset that we do not think that the goal should be to identify a single champion approximate inference technique. Rather, different methods exploit different structural features of large-scale probability models, and we expect that optimal solutions will involve a combination of methods. We return to this point in the discussion section, where we consider various promising hybrids of approximate and exact inference algorithms.

The general problem of approximate inference is NP-hard (Dagum & Luby, 1993) and this provides additional reason to doubt the existence of a single champion approximate inference technique. We think it important to stress, however, that this hardness result, together with Cooper's (1990) hardness result for exact inference cited above, should not be taken to suggest that exact inference and approximate inference are "equally hard." To take an example from a related field, there exist large domains of solid and fluid mechanics in which exact solutions are infeasible but in which approximate techniques (finite element methods) work well. Similarly, in statistical physics, very few models are exactly solvable, but there exist approximate methods (mean field methods, renormalization group methods) that work well in many cases. We feel that the goal of research in probabilistic inference should similarly be that of identifying effective approximate techniques that work well in large classes of problems.

## 2. The QMR-DT Network

The QMR-DT network (Shwe et al., 1991) is a two-level or bi-partite graphical model (see Figure 1). The top level of the graph contains nodes for the *diseases*, and the bottom level contains nodes for the *findings*.

There are a number of conditional independence assumptions reflected in the bi-partite graphical structure. In particular, the diseases are assumed to be marginally independent. (I.e., they are independent in the absence of findings. Note that diseases are *not* assumed to be mutually exclusive; a patient can have multiple diseases). Also, given the states of the disease nodes, the findings are assumed to be conditionally independent. (For a discussion regarding the medical validity and the diagnostic consequences of these and other assumptions embedded into the QMR-DT belief network, see Shwe et al., 1991).

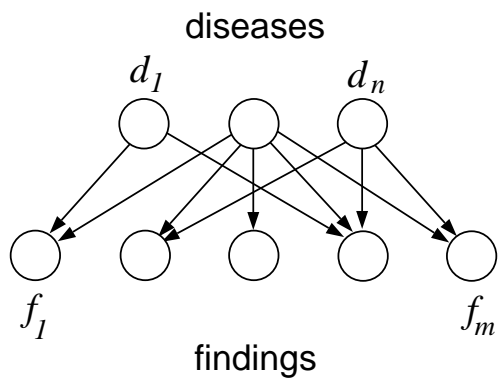

Figure 1: The QMR belief network is a two-level graph where the dependencies between the diseases and their associated findings have been modeled via noisy-OR gates.

To state more precisely the probability model implied by the QMR-DT model, we write the joint probability of diseases and findings as:

$$P(f, d) \;=\; P(f|d)P(d) = \left[\prod_i P(f_i|d)\right]\left[\prod_j P(d_j)\right] \qquad (1)$$

where $d$ and $f$ are binary (1/0) vectors referring to presence/absence states of the diseases and the positive/negative states or outcomes of the findings, respectively. The conditional probabilities $P(f_i|d)$ are represented by the "noisy-OR model" (Pearl, 1988):

$$P(f_i = 0|d) \;=\; P(f_i = 0|L) \prod_{j \in \pi_i} P(f_i = 0|d_j) \qquad (2)$$

$$=\; (1 - q_{i0}) \prod_{j \in \pi_i} (1 - q_{ij})^{d_j} \qquad (3)$$

$$\equiv\; e^{-\theta_{i0} - \sum_{j \in \pi_i} \theta_{ij} d_j}, \qquad (4)$$

where $\pi_i$ is the set of diseases that are parents of the finding $f_i$ in the QMR graph, $q_{ij} = P(f_i = 1|d_j = 1)$ is the probability that the disease $j$, if present, could alone cause the finding to have a positive outcome, and $q_{i0} = P(f_i = 1|L)$ is the "leak" probability, i.e., the probability that the finding is caused by means other than the diseases included in the QMR model. In the final line, we reparameterize the noisy-OR probability model using an exponentiated notation. In this notation, the model parameters are given by $\theta_{ij} = -\log(1 - q_{ij})$.

## 3. Inference

Carrying out diagnostic inference in the QMR model involves computing the posterior marginal probabilities of the diseases given a set of observed positive ($f_i = 1$) and negative ($f_{i'} = 0$) findings. Note that the set of observed findings is considerably smaller than the set of possible findings; note moreover (from the bi-partite structure of the QMR-DT graph) that unobserved findings have no effect on the posterior probabilities for the diseases. For brevity we adopt a notation in which $f_i^+$ corresponds to the event $f_i = 1$, and $f_i^-$ refers to $f_i = 0$ (positive and negative findings respectively). Thus the posterior probabilities of interest are $P(d_j | f^+, f^-)$, where $f^+$ and $f^-$ are the vectors of positive and negative findings.

The negative findings $f^-$ are benign with respect to the inference problem—they can be incorporated into the posterior probability in linear time in the number of associated diseases and in the number of negative findings. As we discuss below, this can be seen from the fact that the probability of a negative finding in Eq. (4) is the exponential of an expression that is linear in the $d_j$. The positive findings, on the other hand, are more problematic. In the worst case the exact calculation of posterior probabilities is exponentially costly in the number of positive findings (Heckerman, 1989; D'Ambrosio, 1994). Moreover, in practical diagnostic situations the number of positive findings often exceeds the feasible limit for exact calculations.

Let us consider the inference calculations in more detail. To find the posterior probability $P(d|f^+, f^-)$, we first absorb the evidence from negative findings, i.e., we compute $P(d|f^-)$. This is just $P(f^-|d)P(d)$ with normalization. Since both $P(f^-|d)$ and $P(d)$ factorize over the diseases (see Eq. (1) and Eq. (2) above), the posterior $P(d|f^-)$ must factorize as well. The normalization of $P(f^-|d)P(d)$ therefore reduces to independent normalizations over each disease and can be carried out in time linear in the number of diseases (or negative findings). In the remainder of the paper, we concentrate solely on the positive findings as they pose the real computational challenge. Unless otherwise stated, we assume that the prior distribution over the diseases already contains the evidence from the negative findings. In other words, we presume that the updates $P(d_j) \leftarrow P(d_j | f^-)$ have already been made.

We now turn to the question of computing $P(d_j | f^+)$, the posterior marginal probability based on the positive findings. Formally, obtaining such a posterior involves marginalizing $P(f^+|d)P(d)$ across the remaining diseases:

$$P(d_j | f^+) \propto \sum_{d \setminus d_j} P(f^+|d)P(d) \qquad (5)$$

where the summation is over all the possible configurations of the disease variables other than $d_j$ (we use the shorthand summation index $d \setminus d_j$ for this). In the QMR model $P(f^+|d)P(d)$ has the form:

$$P(f^+|d)P(d) = \left[ \prod_i P(f_i^+|d) \right] \left[ \prod_j P(d_j) \right] \qquad (6)$$

$$= \left[ \prod_i \left( 1 - e^{-\theta_{i0} - \sum_j \theta_{ij} d_j} \right) \right] \left[ \prod_j P(d_j) \right] \qquad (7)$$

which follows from Eq. (4) and the fact that $P(f_i^+|d) = 1 - P(f^-|d)$. To perform the summation in Eq. (5) over the diseases, we would have to multiply out the terms $1 - e^{\{\cdot\}}$ corresponding to the conditional probabilities for each positive finding. The number of such terms is exponential in the number of positive findings. While algorithms exist that attempt to find and exploit factorizations in this expression, based on the particular pattern of observed evidence (cf. Heckerman, 1989; D'Ambrosio, 1994), these algorithms are limited to roughly 20 positive findings on current computers. It seems unlikely that there is sufficient latent factorization in the QMR-DT model to be able to handle the full CPC corpus, which has a median number of 36 positive findings per case and a maximum number of 61 positive findings.

## 4. Variational Methods

Exact inference algorithms perform many millions of arithmetic operations when applied to complex graphical models such as the QMR-DT. While this proliferation of terms expresses the symbolic structure of the model, it does not necessarily express the numeric structure of the model. In particular, many of the sums in the QMR-DT inference problem are sums over large numbers of random variables. Laws of large numbers suggest that these sums may yield predictable numerical results over the ensemble of their summands, and this fact might enable us to avoid performing the sums explicitly.

To exploit the possibility of numerical regularity in dense graphical models we develop a variational approach to approximate probabilistic inference. Variational methods are a general class of approximation techniques with wide application throughout applied mathematics. Variational methods are particularly useful when applied to highly-coupled systems. By introducing additional parameters, known as "variational parameters"—which essentially serve as low-dimensional surrogates for the high-dimensional couplings of the system—these methods achieve a decoupling of the system. The mathematical machinery of the variational approach provides algorithms for finding values of the variational parameters such that the decoupled system is a good approximation to the original coupled system.

In the case of probabilistic graphical models variational methods allow us to simplify a complicated joint distribution such as the one in Eq. (7). This is achieved via parameterized transformations of the individual node probabilities. As we will see later, these node transformations can be interpreted as delinking the nodes from the graph.

How do we find appropriate transformations? The variational methods that we consider here come from convex analysis (see Appendix 6). Let us begin by considering methods for obtaining upper bounds on probabilities. A well-known fact from convex analysis is that any concave function can be represented as the solution to a minimization problem:

$$f(x) = \min_\xi \{ \xi^T x - f^*(\xi) \} \tag{8}$$

where $f^*(\xi)$ is the *conjugate function* of $f(x)$. The function $f^*(\xi)$ is itself obtained as the solution to a minimization problem:

$$f^*(\xi) = \min_x \{ \xi^T x - f(x) \}. \tag{9}$$

The formal identity of this pair of minimization problems expresses the "duality" of $f$ and its conjugate $f^*$.

The representation of $f$ in Eq. (8) is known as a *variational transformation*. The parameter $\xi$ is known as a *variational parameter*. If we relax the minimization and fix the the variational parameter to an arbitrary value, we obtain an upper bound:

$$f(x) \leq \xi^T x - f^*(\xi). \tag{10}$$

The bound is better for some values of the variational parameter than for others, and for a particular value of $\xi$ the bound is exact.

We also want to obtain lower bounds on conditional probabilities. A straightforward way to obtain lower bounds is to again appeal to conjugate duality and to express functions in terms of a maximization principle. This representation, however, applies to *convex* functions—in the current paper we require lower bounds for *concave* functions. Our concave functions, however, have a special form that allows us to exploit conjugate duality in a different way. In particular, we require bounds for functions of the form $f(a + \sum_j z_j)$, where $f$ is a concave function, where $z_j$ for $i \in \{1, 2, \ldots, n\}$ are non-negative variables, and where $a$ is a constant. The variables $z_j$ in this expression are effectively coupled—the impact of changing one variable is contingent on the settings of the remaining variables. We can use Jensen's inequality, however, to obtain a lower bound in which the variables are decoupled.[3] In particular:

$$f\left(a + \sum_j z_j\right) = f\left(a + \sum_j q_j \frac{z_j}{q_j}\right) \tag{11}$$

$$\geq \sum_j q_j f\left(a + \frac{z_j}{q_j}\right) \tag{12}$$

where the $q_j$ can be viewed as defining a probability distribution over the variables $z_j$. The variational parameter in this case is the probability distribution $q$. The optimal setting of this parameter is given by $q_j = z_j / \sum_k z_k$. This is easily verified by substitution into Eq. (12), and demonstrates that the lower bound is tight.

## 4.1 Variational Upper and Lower Bounds for Noisy-OR

Let us now return to the problem of computing the posterior probabilities in the QMR model. Recall that it is the conditional probabilities corresponding to the positive findings that need to be simplified. To this end, we write

$$P(f_i^+|d) = 1 - e^{-\theta_{i0} - \sum_j \theta_{ij} d_j} = e^{\log(1 - e^{-x})} \tag{13}$$

where $x = \theta_{i0} + \sum_j \theta_{ij} d_j$. Consider the exponent $f(x) = \log(1 - e^{-x})$. For noisy-OR, as well as for many other conditional models involving compact representations (e.g., logistic regression), the exponent $f(x)$ is a concave function of $x$. Based on the discussion in the previous section, we know that there must exist a variational upper bound for this function that is linear in $x$:

$$f(x) \leq \xi x - f^*(\xi) \tag{14}$$

---

3. Jensen's inequality, which states that $f(a + \sum_j q_j x_j) \geq \sum_j q_j f(a + x_j)$, for concave $f$, where $\sum q_j = 1$, and $0 \leq q_j \leq 1$, is a simple consequence of Eq. (8), where $x$ is taken to be $a + \sum_j q_j x_j$.

Using Eq. (9) to evaluate the conjugate function $f^*(\xi)$ for noisy-OR, we obtain:

$$f^*(\xi) = -\xi \log \xi + (\xi + 1) \log(\xi + 1) \tag{15}$$

The desired bound is obtained by substituting into Eq. (13) (and recalling the definition $x = \theta_{i0} + \sum_j \theta_{ij} d_j$):

$$
\begin{align}
P(f_i^+|d) &= e^{f(\theta_{i0} + \sum_j \theta_{ij} d_j)} \tag{16} \\
&\leq e^{\xi_i(\theta_{i0} + \sum_j \theta_{ij} d_j) - f^*(\xi_i)} \tag{17} \\
&\equiv P(f_i^+|d, \xi_i). \tag{18}
\end{align}
$$

Note that the "variational evidence" $P(f_i^+|d, \xi_i)$ is the exponential of a term that is linear in the disease vector $d$. Just as with the negative findings, this implies that the variational evidence can be incorporated into the posterior in time linear in the number of diseases associated with the finding.

There is also a graphical way to understand the effect of the transformation. We rewrite the variational evidence as follows:

$$
\begin{align}
P(f_i^+|d, \xi_i) &= e^{\xi_i(\theta_{i0} + \sum_j \theta_{ij} d_j) - f^*(\xi_i)} \tag{19} \\
&= e^{\xi_i \theta_{i0} - f^*(\xi_i)} \prod_j \left[ e^{\xi_i \theta_{ij}} \right]^{d_j}. \tag{20}
\end{align}
$$

Note that the first term is a constant, and note moreover that the product is factorized across the diseases. Each of the latter factors can be multiplied with the pre-existing prior on the corresponding disease (possibly itself modulated by factors from the negative evidence). The constant term can be viewed as associated with a delinked finding node $f_i$. Indeed, the effect of the variational transformation is to delink the finding node $f_i$ from the graph, altering the priors of the disease nodes that are connected to that finding node. This graphical perspective will be important for the presentation of our variational algorithm—we will be able to view variational transformations as simplifying the graph until a point at which exact methods can be run.

We now turn to the lower bounds on the conditional probabilities $P(f_i^+|d)$. The exponent $f(\theta_{i0} + \sum_j \theta_{ij} d_j)$ in the exponential representation is of the form to which we applied Jensen's inequality in the previous section. Indeed, since $f$ is concave we need only identify the non-negative variables $z_j$, which in this case are $\theta_{ij} d_j$, and the constant $a$, which is now $\theta_{i0}$. Applying the bound in Eq. (12) we have:

$$
\begin{align}
P(f_i^+|d) &= e^{f(\theta_{i0} + \sum_j \theta_{ij} d_j)} \tag{21} \\
&\geq e^{\sum_j q_{j|i} f\left(\theta_{io} + \frac{\theta_{ij} d_j}{q_{j|i}}\right)} \tag{22}
\end{align}
$$

$$= e^{\sum_j q_{j|i} \left[ d_j f \left( \theta_{io} + \frac{\theta_{ij}}{q_{j|i}} \right) + (1 - d_j) f(\theta_{io}) \right]} \tag{23}$$

$$= e^{\sum_j q_{j|i} d_j \left[ f \left( \theta_{io} + \frac{\theta_{ij}}{q_{j|i}} \right) - f(\theta_{io}) \right] + f(\theta_{io})} \tag{24}$$

$$\equiv P(f_i^+ | d, q_{\cdot|i}) \tag{25}$$

where we have allowed a different variational distribution $q_{\cdot|i}$ for each finding. Note that once again the bound is linear in the exponent. As in the case of the upper bound, this implies that the variational evidence can be incorporated into the posterior distribution in time linear in the number of diseases. Moreover, we can once again view the variational transformation in terms of delinking the finding node $f_i$ from the graph.

## 4.2 Approximate Inference for QMR

In the previous section we described how variational transformations are derived for individual findings in the QMR model; we now discuss how to utilize these transformations in the context of an overall inference algorithm.

Conceptually the overall approach is straightforward. Each transformation involves replacing an exact conditional probability of a finding with a lower bound and an upper bound:

$$P(f_i^+ | d, q_{\cdot|i}) \leq P(f_i^+ | d) \leq P(f_i^+ | d, \xi_i). \tag{26}$$

Given that such transformations can be viewed as delinking the $i$th finding node from the graph, we see that the transformations not only yield bounds, but also yield a simplified graphical structure. We can imagine introducing transformations sequentially until the graph is sparse enough that exact methods become feasible. At that point we stop introducing transformations and run an exact algorithm.

There is a problem with this approach, however. We need to decide at each step which node to transform, and this requires an assessment of the effect on overall accuracy of transforming the node. We might imagine calculating the change in a probability of interest both before and after a given transformation, and choosing to transform that node that yields the least change to our target probability. Unfortunately we are unable to calculate probabilities in the original untransformed graph, and thus we are unable to assess the effect of transforming any one node. We are unable to get the algorithm started.

Suppose instead that we work backwards. That is, we introduce transformations for *all* of the findings, reducing the graph to an entirely decoupled set of nodes. We optimize the variational parameters for this fully transformed graph (more on optimization of the variational parameters below). For this graph inference is trivial. Moreover, it is also easy to calculate the effect of reinstating a single exact conditional at one node: we choose to reinstate that node which yields the most change.

Consider in particular the case of the upper bounds (lower bounds are analogous). Each transformation introduces an upper bound on a conditional probability $P(f_i^+ | d)$. Thus the likelihood of observing the (positive) findings $P(f^+)$ is also upper bounded by its variational counterpart $P(f^+ | \xi)$:

$$P(f^+) = \sum_d P(f^+ | d) P(d) \leq \sum_d P(f^+ | d, \xi) P(d) \equiv P(f^+ | \xi) \tag{27}$$

We can assess the accuracy of each variational transformation after introducing and optimizing the variational transformations for all the positive findings. Separately for each positive finding we replace the variationally transformed conditional probability $P(f_i^+|d, \xi_i)$ with the corresponding exact conditional $P(f_i^+|d)$ and compute the difference between the resulting bounds on the likelihood of the observations:

$$\delta_i = P(f^+|\xi) - P(f^+|\xi \setminus \xi_i) \tag{28}$$

where $P(f^+|\xi \setminus \xi_i)$ is computed without transforming the $i^{th}$ positive finding. The larger the difference $\delta_i$ is, the worse the $i^{th}$ variational transformation is. We should therefore introduce the transformations in the ascending order of $\delta_i$s. Put another way, we should treat exactly (not transform) those conditional probabilities whose $\delta_i$ measure is large.

In practice, an intelligent method for ordering the transformations is critical. Figure 2 compares the calculation of likelihoods based on the $\delta_i$ measure as opposed to a method that chooses the ordering of transformations at random. The plot corresponds to a representative diagnostic case, and shows the upper bounds on the log-likelihoods of the observed findings as a function of the number of conditional probabilities that were left intact (i.e. not transformed). Note that the upper bound must improve (decrease) with fewer transformations. The results are striking—the choice of ordering has a large effect on accuracy (note that the plot is on a log-scale).

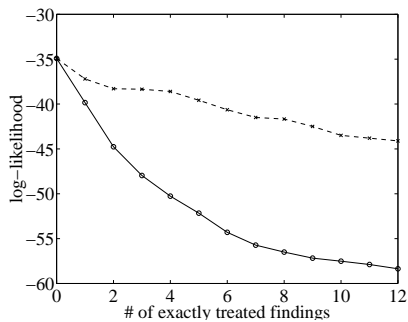

Figure 2: The upper bound on the log-likelihood for the delta method of removing transformations (solid line) and a method that bases the choice on a random ordering (dashed line).

Note also that the curve for the proposed ranking is convex; thus the bound improves less the fewer transformations there are left. This is because we first remove the worst transformations, replacing them with the exact conditionals. The remaining transformations are better as indicated by the delta measure and thus the bound improves less with further replacements.

We make no claims for optimality of the delta method; it is simply a useful heuristic that allows us to choose an ordering for variational transformations in a computationally efficient way. Note also that our implementation of the method optimizes the variational parameters only once at the outset and chooses the ordering of further transformations based on these fixed parameters. These parameters are suboptimal for graphs in which substantial numbers of nodes have been reinstated, but we have found in practice that this simplified algorithm still produces reasonable orderings.

Once we have decided which nodes to reinstate, the approximate inference algorithm can be run. We introduce transformations at those nodes that were left transformed by the ordering algorithm. The product of all of the exact conditional probabilities in the graph with the transformed conditional probabilities yields an upper or lower bound on the overall joint probability associated with the graph (the product of bounds is a bound). Sums of bounds are still bounds, and thus the likelihood (the marginal probability of the findings) is bounded by summing across the bounds on the joint probability. In particular, an upper bound on the likelihood is obtained via:

$$P(f^+) = \sum_d P(f^+|d)P(d) \leq \sum_d P(f^+|d,\xi)P(d) \equiv P(f^+|\xi) \tag{29}$$

and the corresponding lower bound on the likelihood is obtained similarly:

$$P(f^+) = \sum_d P(f^+|d)P(d) \geq \sum_d P(f^+|d,q)P(d) \equiv P(f^+|q) \tag{30}$$

In both cases we assume that the graph has been sufficiently simplified by the variational transformations so that the sums can be performed efficiently.

The expressions in Eq. (29) and Eq. (30) yield upper and lower bounds for arbitrary values of the variational parameters $\xi$ and $q$. We wish to obtain the tightest possible bounds, thus we optimize these expressions with respect to $\xi$ and $q$. We *minimize* with respect to $\xi$ and *maximize* with respect to $q$. Appendix 6 discusses these optimization problems in detail. It turns out that the upper bound is convex in the $\xi$ and thus the adjustment of the variational parameters for the upper bound reduces to a convex optimization problem that can be carried out efficiently and reliably (there are no local minima). For the lower bound it turns out that the maximization can be carried out via the EM algorithm.

Finally, although bounds on the likelihood are useful, our ultimate goal is to approximate the marginal posterior probabilities $P(d_j|f^+)$. There are two basic approaches to utilizing the variational bounds in Eq. (29) and Eq. (30) for this purpose. The first method, which will be our emphasis in the current paper, involves using the transformed probability model (the model based either on upper or lower bounds) as a computationally efficient surrogate for the original probability model. That is, we tune the variational parameters of the transformed model by requiring that the model give the tightest possible bound on the likelihood. We then use the tuned transformed model as an inference engine to provide approximations to other probabilities of interest, in particular the marginal posterior probabilities $P(d_j|f^+)$. The approximations found in this manner are not bounds, but are computationally efficient approximations. We provide empirical data in the following section that show that this approach indeed yields good approximations to the marginal posteriors for the QMR-DT network.

A more ambitious goal is to obtain interval bounds for the marginal posterior probabilities themselves. To this end, let $P(f^+, d_j|\xi)$ denote the combined event that the QMR-DT model generates the observed findings $f^+$ and that the $j^{th}$ disease takes the value $d_j$. These bounds follow directly from:

$$P(f^+, d_j) = \sum_{d \setminus d_j} P(f^+|d)P(d) \leq \sum_{d \setminus d_j} P(f^+|d,\xi)P(d) \equiv P(f^+, d_j|\xi) \tag{31}$$

where $P(f^+|d, \xi)$ is a product of upper-bound transformed conditional probabilities and exact (untransformed) conditionals. Analogously we can compute a *lower* bound $P(f^+, d_j|q)$ by applying the lower bound transformations:

$$P(f^+, d_j) = \sum_{d \backslash d_j} P(f^+|d) P(d) \geq \sum_{d \backslash d_j} P(f^+|d, q) P(d) \equiv P(f^+, d_j|q) \tag{32}$$

Combining these bounds we can obtain interval bounds on the posterior marginal probabilities for the diseases (cf. Draper & Hanks 1994):

$$\frac{P(f^+, d_j|q)}{P(f^+, \bar{d}_j|\xi) + P(f^+, d_j|q)} \leq P(d_j|f^+) \leq \frac{P(f^+, d_j|\xi)}{P(f^+, d_j|\xi) + P(f^+, \bar{d}_j|q)}, \tag{33}$$

where $\bar{d}_j$ is the binary complement of $d_j$.

## 5. Experimental Evaluation

The diagnostic cases that we used in evaluating the performance of the variational techniques were cases abstracted from clinicopathologic conference ("CPC") cases. These cases generally involve multiple diseases and are considered to be clinically difficult cases. They are the cases in which Middleton et al. (1990) did not find their importance sampling method to work satisfactorily.

Our evaluation of the variational methodology consists of three parts. In the first part we exploit the fact that for a subset of the CPC cases (4 of the 48 cases) there are a sufficiently small number of positive findings that we can calculate exact values of the posterior marginals using the Quickscore algorithm. That is, for these four cases we were able to obtain a "gold standard" for comparison. We provide an assessment of the accuracy and efficiency of variational methods on those four CPC cases. We present variational upper and lower bounds on the likelihood as well as scatterplots that compare variational approximations of the posterior marginals to the exact values. We also present comparisons with the likelihood-weighted sampler of Shwe and Cooper (1991).

In the second section we present results for the remaining, intractable CPC cases. We use lengthy runs of the Shwe and Cooper sampling algorithm to provide a surrogate for the gold standard in these cases.

Finally, in the third section we consider the problem of obtaining interval bounds on the posterior marginals.

### 5.1 Comparison to Exact Marginals

Four of the CPC cases have 20 or fewer positive findings (see Table 1), and for these cases it is possible to calculate the exact values of the likelihood and the posterior marginals in a reasonable amount of time. We used Heckerman's "Quickscore" algorithm (Heckerman 1989)—an algorithm tailored to the QMR-DT architecture—to perform these exact calculations.

Figure 3 shows the log-likelihood for the four tractable CPC cases. The figure also shows the variational lower and upper bounds. We calculated the variational bounds twice, with differing numbers of positive findings treated exactly in the two cases ("treated exactly" simply means that the finding is not transformed variationally). In panel (a) there were 8 positive findings treated exactly, and in (b) 12 positive findings were treated exactly. As expected, the bounds were tighter when more positive findings were treated exactly.[4]

| case | # of pos. findings | # of neg. findings |
|------|--------------------|--------------------|
| 1    | 20                 | 14                 |
| 2    | 10                 | 21                 |
| 3    | 19                 | 19                 |
| 4    | 19                 | 33                 |

Table 1: Description of the cases for which we evaluated the exact posterior marginals.

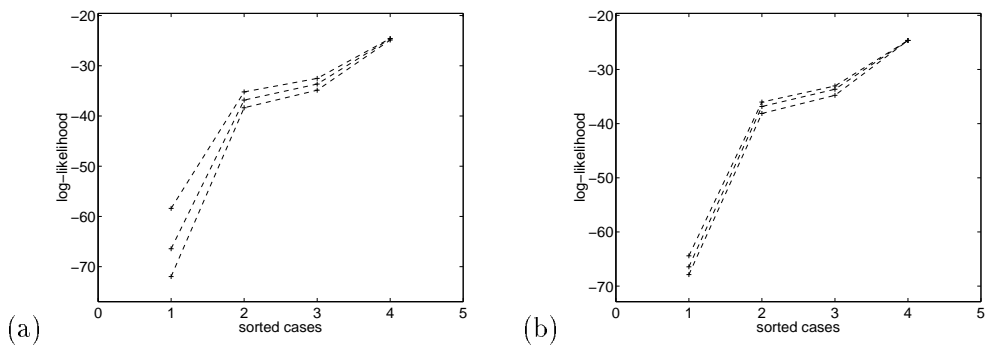

Figure 3: Exact values and variational upper and lower bounds on the log-likelihood $\log P(f^+|\xi)$ for the four tractable CPC cases. In (a) 8 positive findings were treated exactly, and in (b) 12 positive findings were treated exactly.

The average running time across the four tractable CPC cases was 26.9 seconds for the exact method, 0.11 seconds for the variational method with 8 positive findings treated exactly, and 0.85 seconds for the variational method with 12 positive findings treated exactly. (These results were obtained on a 433 MHz DEC Alpha computer).

Although the likelihood is an important quantity to approximate (particularly in applications in which parameters need to be estimated), of more interest in the QMR-DT setting are the posterior marginal probabilities for the individual diseases. As we discussed in the previous section, the simplest approach to obtaining variational estimates of these quantities is to define an approximate variational distribution based either on the distribution $P(f^+|\xi)$, which upper-bounds the likelihood, or the distribution $P(f^+|q)$, which lower-bounds the likelihood. For fixed values of the variational parameters (chosen to provide a tight bound to the likelihood), both distributions provide partially factorized approximations to the joint probability distribution. These factorized forms can be exploited as efficient approximate inference engines for general posterior probabilities, and in particular we can use them to provide approximations to the posterior marginals of individual diseases.

---

4. Given that a significant fraction of the positive findings are being treated exactly in these simulations, one may wonder what if any additional accuracy is due to the variational transformations. We address this concern later in this section and demonstrate that the variational transformations are in fact responsible for a significant portion of the accuracy in these cases.

In practice we found that the distribution $P(f^+|\xi)$ yielded more accurate posterior marginals than the distribution $P(f^+|q)$, and we restrict our presentation to $P(f^+|\xi)$. Figure 4 displays a scatterplot of these approximate posterior marginals, with panel (a) corre-sponding to the case in which 8 positive findings were treated exactly and panel (b) the case in which 12 positive findings treated exactly. The plots were obtained by first extracting the 50 highest posterior marginals from each case using exact methods and then computing the approximate posterior marginals for the corresponding diseases. If the approximate marginals are in fact correct then the points in the figures should align along the diagonals as shown by the dotted lines. We see a reasonably good correspondence—the variational algorithm appears to provide a good approximation to the largest posterior marginals. (We quantify this correspondence with a ranking measure later in this section).

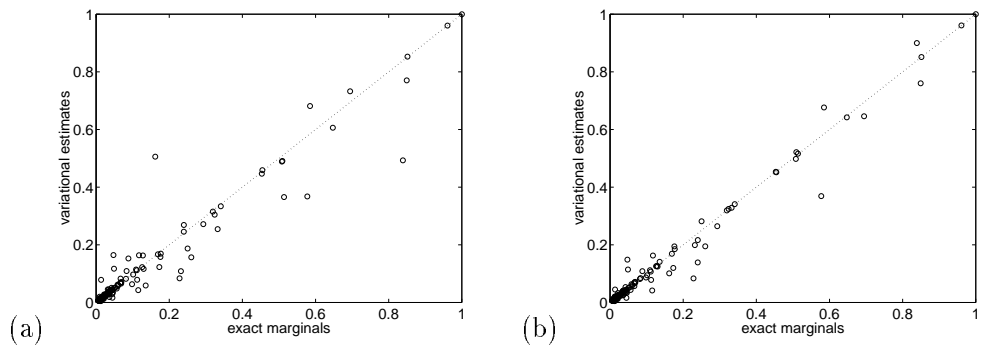

(a)  (b)

Figure 4: Scatterplot of the variational posterior estimates and the exact marginals. In (a) 8 positive findings were treated exactly and in (b) 12 positive findings were treated exactly.

A current state-of-the-art algorithm for the QMR-DT is the enhanced version of likelihood-weighted sampling proposed by Shwe and Cooper (1991). Likelihood-weighted sampling is a stochastic sampling method proposed by Fung and Chang (1990) and Shachter and Peot (1990). Likelihood-weighted sampling is basically a simple forward sampling method that weights samples by their likelihoods. It can be enhanced and improved by utilizing "self-importance sampling" (see Shachter & Peot, 1990), a version of importance sampling in which the importance sampling distribution is continually updated to reflect the current estimated posterior distribution. Middleton et al. (1990) utilized likelihood-weighted sampling with self-importance sampling (as well as a heuristic initialization scheme known as "iterative tabular Bayes") for the QMR-DT model and found that it did not work satisfactorily. Subsequent work by Shwe and Cooper (1991), however, used an additional enhancement to the algorithm known as 'Markov blanket scoring" (see Shachter & Peot, 1990), which distributes fractions of samples to the positive and negative values of a node in proportion to the probability of these values conditioned on the Markov blanket of the node. The combination of Markov blanket scoring and self-importance sampling yielded an effective algorithm.[5] In particular, with these modifications in place, Shwe and Cooper reported reasonable accuracy for two of the difficult CPC cases.

We re-implemented the likelihood-weighted sampling algorithm of Shwe and Cooper, incorporating the Markov blanket scoring heuristic and self-importance sampling. (We did not utilize "iterative tabular Bayes" but instead utilized a related initialization scheme– "heuristic tabular Bayes"–also discussed by Shwe and Cooper). In this section we discuss the results of running this algorithm on the four tractable CPC cases, comparing to the results of variational inference.[6] In the following section we present a fuller comparative analysis of the two algorithms for all of the CPC cases.

Likelihood-weighting sampling, and indeed any sampling algorithm, realizes a time-accuracy tradeoff—taking additional samples requires more time but improves accuracy. In comparing the sampling algorithm to the variational algorithm, we ran the sampling algorithm for several different total time periods, so that the accuracy achieved by the sampling algorithm roughly covered the range achieved by the variational algorithm. The results are shown in Figure 5, with the right-hand curve corresponding to the sampling runs. The figure displays the mean correlations between the approximate and exact posterior marginals across ten independent runs of the algorithm (for the four tractable CPC cases).

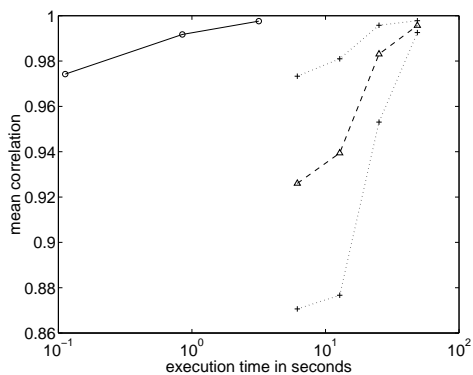

Figure 5: The mean correlation between the approximate and exact posterior marginals as a function of the execution time (in seconds). Solid line: variational estimates; dashed line: likelihood-weighting sampling. The lines above and below the sampling result represent standard errors of the mean based on the ten independent runs of the sampler.

Variational algorithms are also characterized by a time-accuracy tradeoff. In particular, the accuracy of the method generally improves as more findings are treated exactly, at the cost of additional computation. Figure 5 also shows the results from the variational algorithm (the left-hand curve). The three points on the curve correspond to up to 8, 12 and

---

5. The initialization method proved to have little effect on the inference results.

6. We also investigated Gibbs sampling (Pearl, 1988). The results from Gibbs sampling were not as good as the results from likelihood-weighted sampling, and we report only the latter results in the remainder of the paper.

16 positive findings treated exactly. Note that the variational estimates are deterministic and thus only a single run was made.

The figure shows that to achieve roughly equivalent levels of accuracy, the sampling algorithm requires significantly more computation time than the variational method.

Although scatterplots and correlation measures provide a rough indication of the accuracy of an approximation algorithm, they are deficient in several respects. In particular, in diagnostic practice the interest is in the ability of an algorithm to rank diseases correctly, and to avoid both false positives (diseases that are not in fact significant but are included in the set of highly ranked diseases) and false negatives (significant diseases that are omitted from the set of highly ranked diseases). We defined a ranking measure as follows (see also Middleton et al., 1990). Consider a set of the $N$ highest ranking disease hypotheses, where the ranking is based on the correct posterior marginals. Corresponding to this set of diseases we can find the smallest set of $N'$ approximately ranked diseases that includes the $N$ significant ones. In other words, for any $N$ "true positives" an approximate method produces $N' - N$ "false positives." Plotting false positives as a function of true positives provides a meaningful and useful measure of the accuracy of an approximation scheme. To the extent that a method provides a nearly correct ranking of true positives the plot increases slowly and the area under the curve is small. When a significant disease appears late in the approximate ordering the plot increases rapidly near the true rank of the missed disease and the area under the curve is large.

We also plot the number of "false negatives" in a set of top $N$ highly ranked diseases. False negatives refer to the number of diseases, out of the $N$ highest ranking diseases, that do not appear in the set of $N$ approximately ranked diseases. Note that unlike the previous measure, this measure does not reveal the severity of the misplacements, only their frequency.

With this improved diagnostic measure in hand, let us return to the evaluation of the inference algorithms, beginning with the variational algorithm. Figure 6 provides plots of the false positives (panel a) and false negatives (panel b) against the true positives for the tractable CPC cases. Eight positive findings were treated exactly in the simulation shown in this figure. Figure 7 displays the results when 12 positive finding were treated exactly.

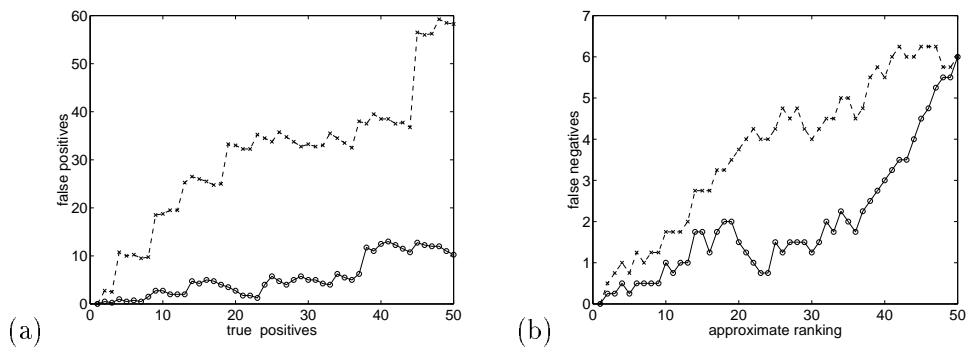

Figure 6: (a) Average number of false positives as a function of true positives for the variational method (solid lines) and the partially-exact method (dashed line). (b) False negatives in the set of top $N$ approximately ranked diseases. In both figures 8 positive findings were treated exactly.

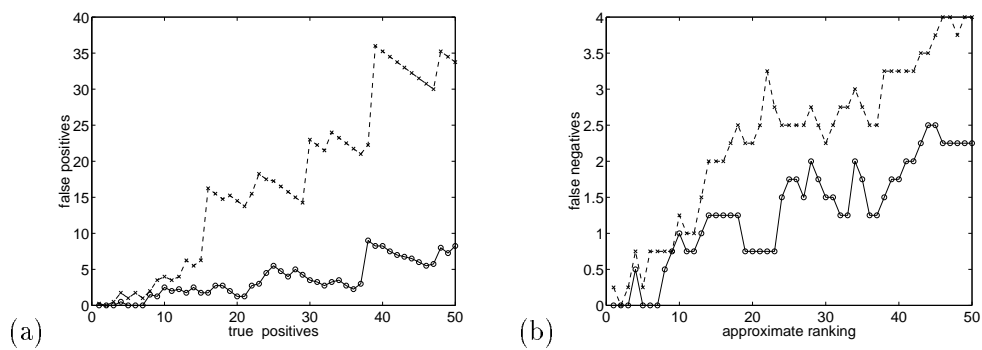

Figure 7: (a) Average number of false positives as a function of true positives for the variational method (solid line) and the partially-exact method (dashed line). (b) False negatives in the set of top $N$ approximately ranked diseases. In both figures 12 positive findings were treated exactly.

As we noted earlier, 8 and 12 positive findings comprise a significant fraction of the total positive findings for the tractable CPC cases, and thus it is important to verify that the variational transformations are in fact contributing to the accuracy of the posterior approximations above and beyond the exact calculations. We did this by comparing the variational method to a method which we call the "partially-exact" method in which the posterior probabilities were obtained using only those findings that were treated exactly in the variational calculations (i.e., using only those findings that were not transformed). If the variational transformations did not contribute to the accuracy of the approximation, then the performance of the partially-exact method should be comparable to that of the variational method.[7] Figure 6 and Figure 7 clearly indicate that this is not the case. The difference in accuracy between these methods is substantial while their computational load is comparable (about 0.1 seconds on a 433MHz Dec Alpha).

We believe that the accuracy portrayed in the false positive plots provides a good indication of the potential of the variational algorithm for providing a practical solution to the approximate inference problem for the QMR-DT. As the figures show, the number of false positives grows slowly with the number of true positives. For example, as shown in Figure 6 where eight positive findings are treated exactly, to find the 20 most likely diseases we would only need to entertain the top 23 diseases in the list of approximately ranked diseases (compared to more than 50 for the partially-exact method).

The ranking plot for the likelihood-weighted sampler is shown in Figure 8, with the curve for the variational method from Figure 7 included for comparison. To make these plots, we ran the likelihood-weighted sampler for an amount of time (6.15 seconds) that was comparable to the time allocated to our slowest variational method (3.17 seconds; this was the case in which 16 positive findings were treated exactly. Recall that the time required for the variational algorithm with 12 positive findings treated exactly was 0.85 seconds.) As the plots show, for these tractable CPC cases, the variational method is significantly more accurate than the sampling algorithm for comparable computational loads.

---

7. It should be noted that this is a conservative comparison, because the partially-exact method in fact benefits from the variational transformation—the set of exactly treated positive findings is selected on the basis of the accuracy of the variational transformations, and these accuracies correlate with the diagnostic relevance of the findings.

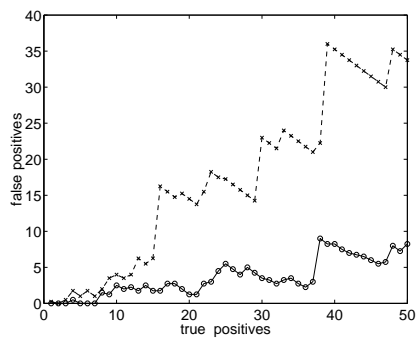

Figure 8: Average number of false positives as a function of true positives for the likelihood-weighted sampler (dashed line) and the variational method (solid line) with 12 positive findings treated exactly.

## 5.2 The Full CPC Corpus

We now consider the full CPC corpus. The majority of these cases (44 of 48 cases), have more than 20 positive findings and thus appear to be beyond the reach of exact methods.

An important attraction of sampling methods is the mathematical guarantee of accurate estimates in the limit of a sufficiently large sample size (Gelfand & Smith, 1990). Thus sampling methods have the promise of providing a general methodology for approximate inference, with two caveats: (1) the number of samples that is needed can be difficult to diagnosis, and (2) very many samples may be required to obtain accurate estimates. For real-time applications, the latter issue can rule out sampling solutions. However, long-term runs of a sampler can still provide a useful baseline for the evaluation of the accuracy of faster approximation algorithms. We begin by considering this latter possibility in the context of likelihood-weighted sampling for the QMR-DT. We then turn to a comparative evaluation of likelihood-weighted sampling and variational methods in the time-limited setting.

To explore the viability of the likelihood-weighted sampler for providing a surrogate for the gold standard, we carried out two independent runs each consisting of 400,000 samples. Figure 9(a) shows the estimates of the log-likelihood from the first sampling run for all of the CPC cases. We also show the variational upper and lower bounds for these cases (the cases have been sorted according to the lower bound). Note that these bounds are rigorous bounds on the true log-likelihood, and thus they provide a direct indication of the accuracy of the sampling estimates. Although we see that many of the estimates lie between the bounds, we also see in many cases that the sampling estimates deviate substantially from the bounds. This suggests that the posterior marginal estimates obtained from these samples are likely to be unreliable as well. Indeed, Figure 9(b) presents a scatterplot of

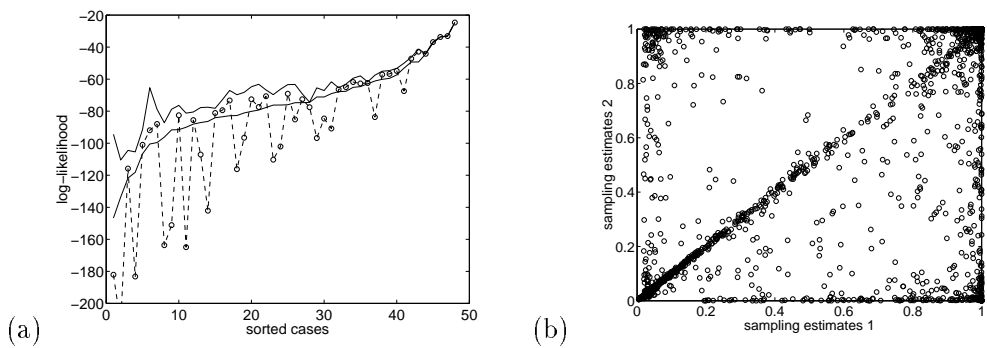

Figure 9: (a) Upper and lower bounds (solid lines) and the corresponding sampling esti-
mates (dashed line) of the log-likelihood of observed findings for the CPC cases.
(b) A correlation plot between the posterior marginal estimates from two inde-
pendent sampling runs.

estimated posterior marginals for the two independent runs of the sampler. Although we
see many cases in which the results lie on the diagonal, indicating agreement between the
two runs, we also see many pairs of posterior estimates that are far from the diagonal.

These results cast some doubt on the viability of the likelihood-weighted sampler as a
general approximator for the full set of CPC cases. Even more problematically we appear
to be without a reliable surrogate for the gold standard for these cases, making it difficult
to evaluate the accuracy of real-time approximations such as the variational method. Note,
however, that the estimates in Figure 9(a) seem to fall into two classes—estimates that
lie within the variational bounds and estimates that are rather far from the bounds. This
suggests the possibility that the distribution being sampled from is multi-modal, with some
estimates falling within the correct mode and providing good approximations and with
others falling in spurious modes and providing seriously inaccurate approximations. If the
situation holds, then an accurate surrogate for the gold standard might be obtained by using
the variational bounds to filter the sampling results and retaining only those estimates that
lie between the bounds given by the variational approach.

Figure 10 provides some evidence of the viability of this approach. In 24 out of the 48
CPC cases both of the independent runs of the sampler resulted in estimates of the log-
likelihood lying approximately within the variational bounds. We recomputed the posterior
marginal estimates for these selected cases and plotted them against each other in the figure.
The scatterplot shows a high degree of correspondence of the posterior estimates in these
cases. We thus tentatively assume that these estimates are accurate enough to serve as a
surrogate gold standard and proceed to evaluate the real-time approximations.

Figure 11 plots the false positives against the true positives on the 24 selected CPC
cases for the variational method. Twelve positive findings were treated exactly in this
simulation. Obtaining the variational estimates took 0.29 seconds of computer time per
case. Although the curve increases more rapidly than with the tractable CPC cases, the
variational algorithm still appears to provide a reasonably accurate ranking of the posterior
marginals, within a reasonable time frame.

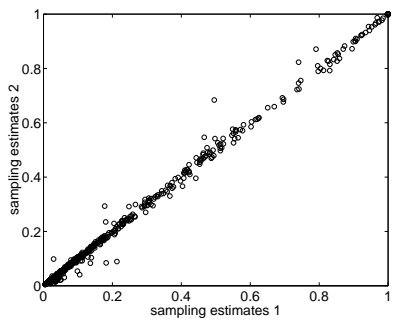

Figure 10: A correlation plot between the selected posterior marginal estimates from two independent sampling runs, where the selection was based on the variational upper and lower bounds.

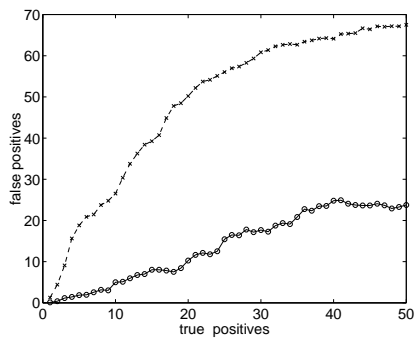

Figure 11: Average number of false positives as a function of true positives for the variational method (solid line) and the likelihood-weighted sampler (dashed line). For the variational method 12 positive findings were treated exactly, and for the sampler the results are averages across ten runs.

To compare the variational algorithm to a time-limited version of the likelihood-weighted sampler we ran the latter algorithm for a period of time (8.83 seconds per case) roughly comparable to the running time of the variational algorithm (0.29 seconds per case). Figure 11 shows the corresponding plot of false positives against true positives, where we have averaged over ten independent runs. We see that the curve increases significantly more steeply than the variational curve. To find the 20 most likely diseases with the variational method we would only need to entertain the top 30 diseases in the list of approximately ranked diseases. For the sampling method we would need to entertain the top 70 approximately ranked diseases.

## 5.3 Interval Bounds on the Marginal Probabilities

Thus far we have utilized the variational approach to produce approximations to the posterior marginals. The approximations that we have discussed originate from upper and lower bounds on the likelihood, but they are not themselves bounds. That is, they are not guaranteed to lie above or below the true posteriors, as we see in Figure 4. As we discussed in Section 4.1, however, it is also possible to induce upper and lower bounds on the posterior marginals from upper and lower bounds on the likelihood (cf. Eq. 33). In this section we evaluate these interval bounds for the QMR-DT posterior marginals.

Figure 12 displays histogram of the interval bounds for the four tractable CPC cases, the 24 selected CPC cases from the previous section, and all of the CPC cases. These histograms include all of the diseases in the QMR-DT network. In the case of the tractable cases the variational method was run with 12 positive findings treated exactly. For the remaining CPC cases the variational method was run with 16 positive findings treated exactly. The running time of the algorithm was less than 10 seconds of computer time per CPC case.

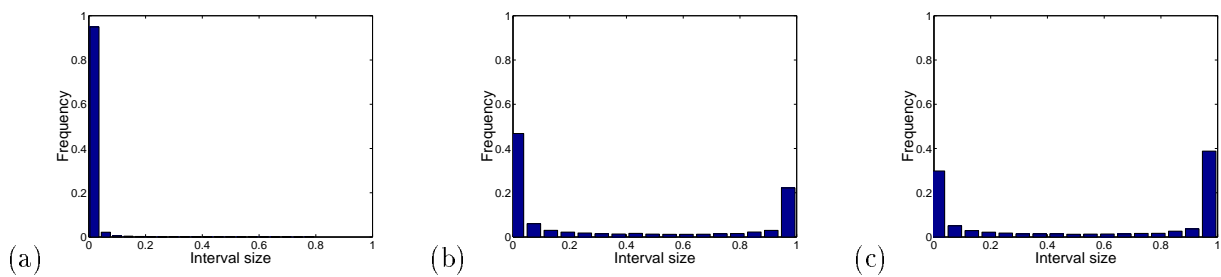

Figure 12: Histograms of the size of the interval bounds on all of the diseases in the QMR-DT network for (a) the four tractable CPC cases, (b) the 24 selected CPC cases from the previous section, and (c) all of the CPC cases.

For the tractable CPC cases, the interval bounds are tight for nearly all of the diseases in the network. However, (1) few of the positive findings are treated variationally in these cases, and (2) there is no need in practice to compute variational bounds for these cases. We get a somewhat better picture of the viability of the variational interval bounds in Figure 12(b) and Figure 12(c), and the picture is decidedly mixed. For the 24 selected cases, tight bounds are provided for approximately half of the diseases. The bounds are vacuous for approximately a quarter of the diseases, and there are a range of diseases in between. When we consider all of the CPC cases, approximately a third of the bounds are tight and nearly half are vacuous.

Although these results may indicate limitations in our variational approximation, there is another more immediate problem that appears to be responsible for the looseness of the bounds in many cases. In particular, recall that we use the Quickscore algorithm (Heckerman, 1989) to handle the exact calculations within the framework of our variational algorithm. Unfortunately Quickscore suffers from vanishing numerical precision for large numbers of positive findings, and in general we begin to run into numerical problems, resulting in vacuous bounds, when 16 positive findings are incorporated exactly into the variational approximation. Thus, although it is clearly of interest to run the variational algorithm for longer durations, and thereby improve the bounds, we are unable to do so within our current implementation of the exact subroutine.

While it is clearly worth studying methods other than Quickscore for treating the exact findings within the variational algorithm, it is also of interest to consider combining variational methods with other methods, such as search-based or other partial evaluation methods, that are based on intervals. These methods may help in simplifying the posterior and obviating the need for improving the exact calculations.

It is also worth emphasizing the positive aspect of these results and their potential practical utility. The previous section showed that the variational method can provide accurate approximations to the posterior marginals. Combined with the interval bounds in this section—which are calculated efficiently—the user can obtain guarantees on approximately a third of these approximations. Given the relatively benign rate of increase in false positives as a function of true positives (Figure 11), such guarantees may suffice. Finally, for diseases in which the bounds are loose there are also perturbation methods available (Jaakkola, 1997) that can help to validate the approximations for these diseases.

## 6. Discussion

Let us summarize the variational inference method and evaluate the results that we have obtained.

The variational method begins with parameterized upper and lower bounds on the individual conditional probabilities at the nodes of the model. For the QMR-DT, these bounds are exponentials of linear functions, and introducing them into the model corresponds to delinking nodes from the graph. Sums of products of these bounds yield bounds, and thus we readily obtain parameterized bounds on marginal probabilities, in particular upper and lower bounds on the likelihood.

We exploited the likelihood bounds in evaluating the output of the likelihood-weighted sampling algorithm. Although the sampling algorithm did not yield reliable results across the corpus of CPC cases, when we utilized the variational upper and lower bounds to select among the samples we were able to obtain sampling results that were consistent between runs. This suggests a general procedure in which variational bounds are used to assess the convergence of a sampling algorithm. (One can also imagine a more intimate relationship between these algorithms in which the variational bounds are used to adjust the on-line course of the sampler).

The fact that we have bounds on the likelihood (or other marginal probabilities) is critical—the bounding property allows us to find optimizing values of the variational parameters by minimizing the upper-bounding variational distribution and maximizing the lower-bounding variational distribution. In the case of the QMR-DT network (a bipartite noisy-OR graph), the minimization problem is a convex optimization problem and the maximization problem is solved via the EM algorithm.

Once the variational parameters are optimized, the resulting variational distribution can be exploited as an inference engine for calculating approximations to posterior probabilities. This technique has been our focus in the paper. Graphically, the variationally transformed model can be viewed as a sub-graph of the original model in which some of the finding nodes have been delinked. If a sufficient number of findings are delinked variationally then it is possible to run an exact algorithm on the resulting graph. This approach yields approximations to the posterior marginals of the disease nodes.

We found empirically that these approximations appeared to provide good approximations to the true posterior marginals. This was the case for the tractable set of CPC cases (cf. Figure 7) and—subject to our assumption that we have obtained a good surrogate for the gold standard via the selected output of the sampler—also the case for the full CPC corpus (cf. Figure 11).

We also compared the variational algorithm to a state-of-the-art algorithm for the QMR-DT, the likelihood-weighted sampler of Shwe and Cooper (1991). We found that the variational algorithm outperformed the likelihood-weighted sampler both for the tractable cases and for the full corpus. In particular, for a fixed accuracy requirement the variational algorithm was significantly faster (cf. Figure 5), and for a fixed time allotment the variational algorithm was significantly more accurate (cf. Figure 8 and Figure 11).

Our results were less satisfactory for the interval bounds on the posterior marginals. Across the full CPC corpus we found that for approximately one third of the disease the bounds were tight but for half of the diseases the bounds were vacuous. A major impediment to obtaining tighter bounds appears to lie not in the variational approximation per se but rather in the exact subroutine, and we are investigating exact methods with improved numerical properties.

Although we have focused in detail on the QMR-DT model in this paper, it is worth noting that the variational probabilistic inference methodology is considerably more general. Specifically, the methods that we have described here are not limited to the bi-partite graphical structure of the QMR-DT model, nor is it necessary to employ noisy-OR nodes (Jaakkola & Jordan, 1996). It is also the case that the type of transformations that we have exploited in the QMR-DT setting extend to a larger class of dependence relations based on generalized linear models (Jaakkola, 1997). Finally, for a review of applications of variational methods to a variety of other graphical model architectures, see Jordan, et al. (1998).

A promising direction for future research appears to be in the integration of various kinds of approximate and exact methods (see, e.g., Dagum & Horvitz, 1992; Jensen, Kong, & Kjærulff, 1995). In particular, search-based methods (Cooper, 1985; Peng & Reggia, 1987, Henrion, 1991) and variational methods both yield bounds on probabilities, and, as we have indicated in the introduction, they seem to exploit different aspects of the structure of complex probability distributions. It may be possible to combine the bounds from these algorithm—the variational bounds might be used to guide the search, or the search-based bounds might be used to aid the variational approximation. Similar comments can be made with respect to localized partial evaluation methods and bounded conditioning methods (Draper & Hanks, 1994; Horvitz, et al., 1989). Also, we have seen that variational bounds can be used for assessing whether estimates from Monte Carlo sampling algorithms have converged. A further interesting hybrid would be a scheme in which variational approximations are refined by treating them as initial conditions for a sampler.

Even without extensions our results in this paper appear quite promising. We have presented an algorithm which runs in real time on a large-scale graphical model for which exact algorithms are in general infeasible. The results that we have obtained appear to be reasonably accurate across a corpus of difficult diagnostic cases. While further work is needed, we believe that our results indicate a promising role for variational inference in developing, critiquing and exploiting large-scale probabilistic models such as the QMR-DT.

## Acknowledgements

We would like to thank the University of Pittsburgh and Randy Miller for the use of the QMR-DT database. We also want to thank David Heckerman for suggesting that we attack QMR-DT with variational methods, and for providing helpful counsel along the way.

## Appendix A. Duality

The upper and lower bounds for individual conditional probability distributions that form the basis of our variational method are based on the "dual" or "conjugate" representations of convex functions. We present a brief description of convex duality in this appendix, and refer the reader to Rockafellar (1970) for a more extensive treatment.

Let $f(x)$ be a real-valued, convex function defined on a convex set $X$ (for example, $X = R^n$). For simplicity of exposition, we assume that $f$ is a well-behaved (differentiable) function. Consider the graph of $f$, i.e., the points $(x, f(x))$ in an $n + 1$ dimensional space. The fact that the function $f$ is convex translates into convexity of the set $\{(x, y) : y \geq f(x)\}$ called the *epigraph* of $f$ and denoted by $epi(f)$ (Figure 13). It is an elementary property

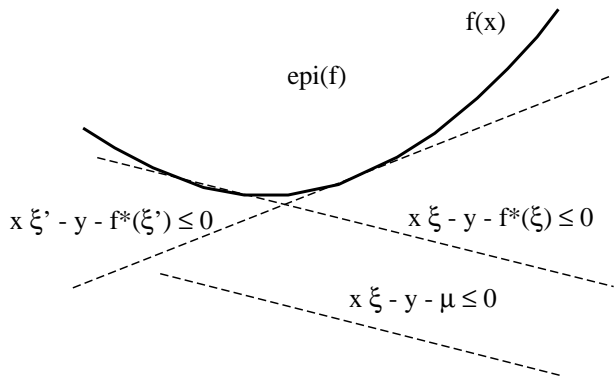

Figure 13: Half-spaces containing the convex set $epi(f)$. The conjugate function $f^*(\xi)$ defines the critical half-spaces whose intersection is $epi(f)$, or, equivalently, it defines the tangent planes of $f(x)$.

of convex sets that they can be represented as the intersection of all the half-spaces that contain them (see Figure 13). Through parameterizing these half-spaces we obtain the dual representations of convex functions. To this end, we define a half-space by the condition:

$$\text{all } (x, y) \text{ such that } x^T \xi - y - \mu \leq 0 \tag{34}$$

where $\xi$ and $\mu$ parameterize all (non-vertical) half-spaces. We are interested in characterizing the half-spaces that contain the epigraph of $f$. We require therefore that the points in the epigraph must satisfy the half-space condition: for $(x, y) \in epi(f)$, we must have $x^T \xi - y - \mu \leq 0$. This holds whenever $x^T \xi - f(x) - \mu \leq 0$ as the points in the epigraph have the property that $y \geq f(x)$. Since the condition must be satisfied by all $x \in X$, it follows

that

$$\max_{x \in X} \{ x^T \xi - f(x) - \mu \} \leq 0, \qquad (35)$$

as well. Equivalently,

$$\mu \quad \geq \quad \max_{x \in X} \{ x^T \xi - f(x) \} \qquad (36)$$

where the right-hand side of this equation defines a function of $\xi$, which is known as the "dual" or "conjugate" function $f^*(\xi)$. This function, which is also a convex function, defines the critical half-spaces which are needed for the representation of $epi(f)$ as an intersection of half-spaces (Figure 13).

To clarify the duality between $f(x)$ and $f^*(x)$, let us drop the maximum and rewrite the inequality as:

$$x^T \xi \leq f(x) + f^*(\xi) \qquad (37)$$

In this equation, the roles of the two functions are interchangeable and we may suspect that also $f(x)$ can obtained from the dual function $f^*(x)$ by an optimization procedure. This is in fact the case and we have:

$$f(x) = \max_{\xi \in \Xi} \{ x^T \xi - f^*(\xi) \} \qquad (38)$$

This equality states that the dual of the dual gives back the original function. It provides the computational tool for calculating dual functions.

For concave (convex down) functions the results are analogous; we replace max with min, and lower bounds with upper bounds.

## Appendix B. Optimization of the Variational Parameters

The variational method that we have described involves replacing selected local conditional probabilities with either upper-bounding or lower-bounding variational transformations. Because the product of bounds is a bound, the variationally transformed joint probability distribution is a bound (upper or lower) on the true joint probability distribution. Moreover, because sums of bounds is a bound on the sum, we can obtain bounds on marginal probabilities by marginalizing the variationally transformed joint probability distribution. In particular, this provides a method for obtaining bounds on the likelihood (the marginal probability of the evidence).

Note that the variationally transformed distributions are bounds for arbitrary values of the variational parameters (because each individually transformed node conditional probability is a bound for arbitrary values of its variational parameter). To obtain optimizing values of the variational parameters, we take advantage of the fact that our transformed distribution is a bound, and either minimize (in the case of upper bounds) or maximize (in the case of lower bounds) the transformed distribution with respect to the variational parameters. It is this optimization process which provides a tight bound on the marginal probability of interest (e.g., the likelihood) and thereby picks out a particular variational distribution that can subsequently be used for approximate inference.

In this appendix we discuss the optimization problems that we must solve in the case of noisy-OR networks. We consider the upper and lower bounds separately, beginning with the upper bound.

## Upper Bound Transformations

Our goal is to compute a tight upper bound on the likelihood of the observed findings: $P(f^+) = \sum_d P(f^+|d)P(d)$. As discussed in Section 4.2, we obtain an upper bound on $P(f^+|d)$ by introducing upper bounds for individual node conditional probabilities. We represent this upper bound as $P(f^+|d, \xi)$, which is a product across the individual variational transformations and may contain contributions due to findings that are being treated exactly (i.e., are not transformed). Marginalizing across $d$ we obtain a bound:

$$P(f^+) \leq \sum_d P(f^+|d, \xi)P(d) \equiv P(f^+|\xi). \tag{39}$$

It is this latter quantity that we wish to minimize with respect to the variational parameters $\xi$.

To simplify the notation we assume that the first $m$ positive findings have been transformed (and therefore need to be optimized) while the remaining conditional probabilities will be treated exactly. In this notation $P(f^+|\xi)$ is given by

$$P(f^+|\xi) = \sum_d \left[ \prod_{i \leq m} P(f_i^+|d, \xi_i) \right] \left[ \prod_{i > m} P(f_i^+|d) \right] \prod_j P(d_j) \tag{40}$$

$$\propto E \left\{ \prod_{i \leq m} P(f_i^+|d, \xi_i) \right\}, \tag{41}$$

where the expectation is taken with respect to the posterior distribution for the diseases given those positive findings that we plan to treat exactly. Note that the proportionality constant does not depend on the variational parameters (it is the likelihood of the exactly treated positive findings). We now insert the explicit forms of the transformed conditional probabilities (see Eq. (17)) into Eq. (41) and find:

$$P(f^+|\xi) \propto E \left\{ \prod_{i \leq m} e^{\xi_i(\theta_{i0} + \sum_j \theta_{ij}d_j) - f^*(\xi_i)} \right\} \tag{42}$$

$$= e^{\sum_{i \leq m} (\xi_i \theta_{i0} - f^*(\xi_i))} E \left\{ e^{\sum_{j,\, i \leq m} \xi_i \theta_{ij}d_j} \right\} \tag{43}$$

where we have simply converted the products over $i$ into sums in the exponent and pulled out the terms that are constants with respect to the expectation. On a log-scale, the proportionality becomes an equivalence up to a constant:

$$\log P(f^+|\xi) = C + \sum_{i \leq m} (\xi_i \theta_{i0} - f^*(\xi_i)) + \log E \left\{ e^{\sum_{j,\, i \leq m} \xi_i \theta_{ij}d_j} \right\} \tag{44}$$

Several observations are in order. Recall that $f^*(\xi_i)$ is the conjugate of the concave function $f$ (the exponent), and is therefore also concave; for this reason $-f^*(\xi_i)$ is convex. In Appendix C we prove that the remaining term:

$$\log E\left\{e^{\sum_{j,i\leq m}\xi_i\theta_{ij}d_j}\right\} \tag{45}$$

is also a convex function of the variational parameters. Now, since any sum of convex functions is convex, we conclude that $\log P(f^+|\xi)$ is a convex function of the variational parameters. This means that there are no local minima in our optimization problem. We may safely employ the standard Newton-Raphson procedure to solve $\nabla \log P(f^+|\xi) = 0$. Alternatively we can utilize fixed-point iterations. In particular, we calculate the derivatives of the variational form and iteratively solve for the individual variational parameters $\xi_k$ such that the derivatives are zero. The derivatives are given as follows:

$$\frac{\partial}{\partial \xi_k}\log P(f^+|\xi) = \theta_{k0} + \log\frac{\xi_k}{1+\xi_k} + E\left\{\sum_j \theta_{kj}d_j\right\} \tag{46}$$

$$\frac{\partial^2}{\partial^2 \xi_k}\log P(f^+|\xi) = \frac{1}{\xi_k} - \frac{1}{1+\xi_k} + Var\left\{\sum_j \theta_{kj}d_j\right\}, \tag{47}$$

where the expectation and the variance are with respect to the posterior approximation $P(d|f^+,\xi)$, and both derivatives can be computed in time linear in the number of associated diseases for the finding. The benign scaling of the variance calculations comes from exploiting the special properties of the noisy-OR dependence and the marginal independence of the diseases.

Calculating the expectations in Eq. (7) is exponentially costly in the number of exactly treated positive findings. When there are a large number of positive findings, we can have recourse to a simplified procedure in which we optimize variational parameters after having transformed all or most of the positive findings. While the resulting variational parameters are suboptimal, we have found in practice that the incurred loss in accuracy is typically quite small. In the simulations reported in the paper, we optimized the variational parameters after approximately half of the exactly treated findings had been introduced. (To be precise, in the case of 8, 12 and 16 total findings treated exactly, we optimized the parameters after 4, 8, and 8 findings, respectively, were introduced).

## Lower Bound Transformations

Mimicking the case of upper bounds, we replace individual conditional probabilities of the findings with lower-bounding transformations, resulting in a lower-bounding expression $P(f^+|d,q)$. Taking the product with $P(d)$ and marginalizing over $d$ yields a lower bound on the likelihood:

$$P(f^+) \geq \sum_d P(f^+|d,q)P(d) \equiv P(f^+|q). \tag{48}$$

We wish to maximize $P(f^+|q)$ with respect to the variational parameters $q$ to obtain the tightest possible bound.

Our problem can be mapped onto a standard optimization problem in statistics. In particular, treating $d$ as a latent variable, $f$ as an observed variable, and $q$ as a parameter vector, the optimization of $P(f^+|q)$ (or its logarithm) can be viewed as a standard maximum likelihood estimation problem for a latent variable model. It can be solved using the EM algorithm (Dempster, Laird, & Rubin, 1977). The algorithm yields a sequence of variational parameters that monotonically increase the objective function $\log P(f^+|q)$. Within the EM framework, we obtain an update of the variational parameters by maximizing the expected complete log-likelihood:

$$E\left\{\log P(f^+|d,q)P(d)\right\} = \sum_i E\left\{\log P(f_i^+|d,q_{\cdot|i})\right\} + \text{constant},\qquad(49)$$

where $q^{old}$ denotes the vector of variational parameters before the update, where the constant term is independent of the variational parameters $q$ and where the expectation is with respect to the posterior distribution $P(d|f^+,q^{old}) \propto P(f^+|d,q^{old})P(d)$. Since the variational parameters associated with the conditional probabilities $P(f_i^+|d,q_{\cdot|i})$ are independent of one another, we can maximize each term in the above sum separately. Recalling the form of the variational transformation (see Eq. (24)), we have:

$$E\left\{\log P(f_i^+|d,q_{\cdot|i})\right\} = \sum_j q_{j|i}\, E\{d_j\}\left[f\left(\theta_{io} + \frac{\theta_{ij}}{q_{j|i}}\right) - f(\theta_{io})\right]$$
$$+ f(\theta_{io})\qquad(50)$$

which we are to maximize with respect to $q_{j|i}$ while keeping the expectations $E\{d_j\}$ fixed. This optimization problem can be solved iteratively and monotonically by performing the following synchronous updates with normalization:

$$q_{j|i} \;\leftarrow\; E\{d_j\}\left[q_{j|i}\, f\left(\theta_{io} + \frac{\theta_{ij}}{q_{j|i}}\right) - \theta_{ij}\, f'\left(\theta_{io} + \frac{\theta_{ij}}{q_{j|i}}\right) - q_{j|i}\, f(\theta_{io})\right]\qquad(51)$$

where $f'$ denotes the derivative of $f$. (The update is guaranteed to be non-negative).

This algorithm can be easily extended to handle the case where not all the positive findings have been transformed. The only new feature is that some of the conditional probabilities in the products $P(f^+|d,q^{old})$ and $P(f^+|d,q)$ have been left intact, i.e., not transformed; the optimization with respect to the variational parameters corresponding to the transformed conditionals proceeds as before.

## Appendix C. Convexity

The purpose of this appendix is to demonstrate that the function:

$$\log E\left\{e^{\sum_{j,i\leq m}\xi_i\theta_{ij}d_j}\right\}\qquad(52)$$

is a convex function of the variational parameters $\xi_i$. We note first that affine transformations do not change convexity properties. Thus convexity in $X = \sum_{j,i\leq m}\xi_i\theta_{ij}d_j$ implies convexity in the variational parameters $\xi$. It remains to show that

$$\log E\left\{e^{X}\right\} = \log \sum_i p_i\, e^{X_i} = f(\vec{X}) \tag{53}$$

is a convex function of the vector $\vec{X} = \{X_1 \ldots X_n\}^T$; here we have indicated the discrete values in the range of the random variable $X$ by $X_i$ and denoted the probability measure on such values by $p_i$. Taking the gradient of $f$ with respect to $X_k$ gives:

$$\frac{\partial}{\partial X_k} f(\vec{X}) = \frac{p_k e^{X_k}}{\sum_i p_i\, e^{X_i}} \equiv q_k \tag{54}$$

where $q_k$ defines a probability distribution. The convexity is revealed by a positive semi-definite Hessian $\mathcal{H}$, whose components in this case are

$$\mathcal{H}_{kl} = \frac{\partial^2}{\partial X_k \partial X_l} f(\vec{X}) = \delta_{kl} q_k - q_k q_l \tag{55}$$

To see that $\mathcal{H}$ is positive semi-definite, consider

$$\vec{Z}^T \mathcal{H} \vec{Z} = \sum_k q_k Z_k^2 - (\sum_k q_k Z_k)(\sum_l q_l Z_l) = \mathrm{Var}\{Z\} \geq 0 \tag{56}$$

where $\mathrm{Var}\{Z\}$ is the variance of a discrete random variable $Z$ which takes the values $Z_i$ with probability $q_i$.

## References

D'Ambrosio, B. (1993). Incremental probabilistic inference. In *Proceedings of the Ninth Conference on Uncertainty in Artificial Intelligence*. San Mateo, CA: Morgan Kaufmann.

D'Ambrosio, B. (1994). Symbolic probabilistic inference in large BN20 networks. In *Proceedings of the Tenth Conference on Uncertainty in Artificial Intelligence*. San Mateo, CA: Morgan Kaufmann.

Cooper, G. (1985). *NESTOR: A computer-based medical diagnostic aid that integrates causal and probabilistic knowledge*. Ph.D. Dissertation, Medical Informatics Sciences, Stanford University, Stanford, CA. (Available from UMI at http://wwwlib.umi.com/dissertations/main).

Cooper, G. (1990). The computational complexity of probabilistic inference using Bayesian belief networks. *Artificial Intelligence, 42*, 393–405.

Dagum, P., & Horvitz, E. (1992). Reformulating inference problems through selective conditioning. In *Proceedings of the Eighth Annual Conference on Uncertainty in Artificial Intelligence*.

Dagum, P., & Horvitz, E. (1993). A Bayesian analysis of simulation algorithms for inference in Belief networks. *Networks, 23*, 499–516.

Dagum, P., & Luby, M. (1993). Approximate probabilistic reasoning in Bayesian belief networks is NP-hard. *Artificial Intelligence, 60*, 141–153.

Dechter, R. (1997). Mini-buckets: A general scheme of generating approximations in automated reasoning. In *Proceedings of the Fifteenth International Joint Conference on Artificial Intelligence.*

Dechter, R. (1998). Bucket elimination: A unifying framework for probabilistic inference. In M. I. Jordan (Ed.), *Learning in Graphical Models.* Cambridge, MA: MIT Press.

Dempster, A., Laird, N., & Rubin, D. (1977). Maximum likelihood from incomplete data via the EM algorithm. *Journal of the Royal Statistical Society B, 39*, 1–38.

Draper, D., & Hanks, S. (1994). Localized partial evaluation of belief networks. In *Proceedings of the Tenth Annual Conference on Uncertainty in Artificial Intelligence.*

Fung, R., & Chang, K. C. (1990). Weighting and integrating evidence for stochastic simulation in Bayesian networks. In *Proceedings of Fifth Conference on Uncertainty in Artificial Intelligence.* Amsterdam: Elsevier Science.

Gelfand, A., & Smith, A. (1990). Sampling-based approaches to calculating marginal Densities. *Journal of the American Statistical Association, 85*, 398–409.

Heckerman, D. (1989). A tractable inference algorithm for diagnosing multiple diseases. In *Proceedings of the Fifth Conference on Uncertainty in Artificial Intelligence.*

Henrion, M. (1991). Search-based methods to bound diagnostic probabilities in very large belief nets. In *Proceedings of Seventh Conference on Uncertainty in Artificial Intelligence.*

Horvitz, E. Suermondt, H., & Cooper, G. (1989). Bounded conditioning: Flexible inference for decisions under scarce resources. In *Proceedings of Fifth Conference on Uncertainty in Artificial Intelligence.*

Jaakkola, T. (1997). *Variational methods for inference and learning in graphical models.* PhD thesis, Department of Brain and Cognitive Sciences, Massachusetts Institute of Technology.

Jaakkola, T., & Jordan, M. (1996). Recursive algorithms for approximating probabilities in graphical models. In *Advances of Neural Information Processing Systems 9.* Cambridge, MA: MIT Press.

Jensen, C. S., Kong, A., & Kjærulff, U. (1995). Blocking-Gibbs sampling in very large probabilistic expert systems. *International Journal of Human-Computer Studies, 42*, 647–666.

Jensen, F. (1996). *Introduction to Bayesian networks.* New York: Springer.

Jordan, M., Ghaharamani, Z. Jaakkola, T., & Saul, L. (in press). An introduction to variational methods for graphical models. *Machine Learning*.

Lauritzen, S., & Spiegelhalter, D. (1988). Local computations with probabilities on graphical structures and their application to expert systems (with discussion). *Journal of the Royal Statistical Society B, 50*, 157–224.

MacKay, D. J. C. (1998). Introduction to Monte Carlo methods. In M. I. Jordan (Ed.), *Learning in Graphical Models*. Cambridge, MA: MIT Press.

Middleton, B., Shwe, M., Heckerman, D., Henrion, M., Horvitz, E., Lehmann, H., & Cooper, G. (1990). Probabilistic diagnosis using a reformulation of the INTERNIST-1/QMR knowledge base II. Evaluation of diagnostic performance. Section on Medical Informatics Technical report SMI-90-0329, Stanford University.

Miller, R. A., Fasarie, F. E., & Myers, J. D. (1986). Quick medical reference (QMR) for diagnostic assistance. *Medical Computing, 3*, 34–48.

Pearl, J. (1988). *Probabilistic Reasoning in Intelligent Systems*. San Mateo, CA: Morgan Kaufmann.

Peng, Y., & Reggia, J. (1987). A probabilistic causal model for diagnostic problem solving – Part 2: Diagnostic strategy. *IEEE Trans. on Systems, Man, and Cybernetics: Special Issue for Diagnosis, 17*, 395–406.

Poole, D. (1997). Probabilistic partial evaluation: Exploiting rule structure in probabilistic inference. In *Proceedings of the Fifteenth International Joint Conference on Artificial Intelligence*.

Rockafellar, R. (1972). *Convex Analysis*. Princeton University Press.

Shachter, R. D., & Peot, M. (1990). Simulation approaches to general probabilistic inference on belief networks. In *Proceedings of Fifth Conference on Uncertainty in Artificial Intelligence*. Elsevier Science: Amsterdam.

Shenoy, P. P. (1992). Valuation-based systems for Bayesian decision analysis. *Operations Research, 40*, 463–484.

Shwe, M., & Cooper, G. (1991). An empirical analysis of likelihood – weighting simulation on a large, multiply connected medical belief network. *Computers and Biomedical Research, 24*, 453-475.

Shwe, M., Middleton, B., Heckerman, D., Henrion, M., Horvitz, E., Lehmann, H., & G. Cooper (1991). Probabilistic diagnosis using a reformulation of the INTERNIST-1/QMR knowledge base I. The probabilistic model and inference algorithms. *Methods of Information in Medicine, 30*, 241–255.